\documentclass{article}

\usepackage{microtype}
\usepackage{graphicx}
\usepackage{subfigure}
\usepackage{booktabs} %

\usepackage{hyperref}
\usepackage{placeins} %

\usepackage[accepted]{icml2025}

\usepackage{mathtools}
\usepackage{amsthm}

\usepackage{multirow}
\usepackage{hyperref}       %
\usepackage{url}            %
\usepackage{booktabs}       %
\usepackage{amsfonts}       %
\usepackage{nicefrac}       %
\usepackage{microtype}      %
\usepackage{xcolor}         %
\usepackage{multicol}
\usepackage{amsmath}
\usepackage{bbm}
\usepackage{graphicx}

\usepackage{wrapfig}
\usepackage{placeins}
\usepackage{subcaption}
\usepackage{caption}
\usepackage{array}
\usepackage{amssymb}
\usepackage{mathrsfs}
\usepackage{algorithm}
\usepackage{algorithmic}
\usepackage{xspace}

\usepackage[capitalize,noabbrev]{cleveref}

\usepackage{soul} %
\usepackage{xcolor} %
\definecolor{verylightgray}{rgb}{0.9, 0.9, 0.9}
\sethlcolor{verylightgray} %
\definecolor{verylightblue}{rgb}{0.9, 0.95, 1.0} %
\definecolor{verylightpurple}{rgb}{0.95, 0.9, 1.0} %

\newcommand{\hlgray}[1]{\sethlcolor{verylightgray}\hl{#1}}
\newcommand{\hlblue}[1]{\sethlcolor{verylightblue}\hl{#1}} %
\newcommand{\hlpurple}[1]{\sethlcolor{verylightpurple}\hl{#1}} %

\theoremstyle{plain}

\theoremstyle{definition}

\theoremstyle{remark}

\usepackage[textsize=tiny]{todonotes}

\begin{document}

\twocolumn[
\icmltitle{In-context KV-Cache Eviction for LLMs via Attention-Gate}

\begin{icmlauthorlist}
\icmlauthor{Zihao Zeng}{sjtu}
\icmlauthor{Bokai Lin}{sjtu}
\icmlauthor{Tianqi Hou}{huawei}
\icmlauthor{Hao Zhang}{ucsd}
\icmlauthor{Zhijie Deng}{sjtu}
\end{icmlauthorlist}

\icmlaffiliation{sjtu}{Shanghai Jiao Tong University}
\icmlaffiliation{huawei}{Huawei}
\icmlaffiliation{ucsd}{University of California, San Diego}

\icmlcorrespondingauthor{Zhijie Deng}{zhijied@sjtu.edu.cn}

\icmlkeywords{Machine Learning, ICML}

\vskip 0.3in
]

\printAffiliationsAndNotice{}  %

\begin{abstract}
The KV-Cache technique has become the standard for the inference of large language models (LLMs).
Yet, it is widely criticized that KV-Cache can become a bottleneck of the LLM inference system.
This paper enables a novel dynamic KV-Cache eviction policy by injecting a lightweight module called \emph{Attention-Gate} to the model.
It accepts the \emph{global} context as input and yields eviction flags for each token.
The self-attention modules in the model proceed according to the flags and cache only a subset of the KV states for next token prediction.
The Attention-Gates can yield various flags for different heads and layers and be easily tuned on top of a pre-trained LLM via continual pre-training or supervised fine-tuning.
The computational and memory overhead introduced by Attention-Gates can be minimal.
We empirically evaluate the proposed approach across multiple scenarios, showing that effective eviction of redundant tokens can not only improve efficiency but also enhance performance.

\end{abstract}

\section{Introduction}

Large language models (LLMs)~\citep{dubey2024llama, team2024gemma, vicuna2023} have achieved remarkable success across a wide range of tasks.
A key technique that has enabled efficient LLM inference is KV-Cache, which stores transient attention keys and values to avoid recomputation.
However, as the size of LLMs continues to increase and the demand for handling long-context queries grows, the KV-Cache has emerged as a significant bottleneck.
Storing attention states for numerous tokens can lead to considerable memory overhead and data transfer among the memory hierarchy results in substantially increased inference time.

Studies have shown that sparsity is a natural phenomenon in attention mechanisms, with many tokens being redundant for inference~\citep{zhang2024h2o}.
This suggests that retaining all tokens in the KV-Cache is unnecessary.
Existing works have explored this insight
to compress KV-Cache using static strategies or hinging on accumulative attention scores.
StreamingLLM~\citep{xiao2024efficientstreaminglanguagemodels} is a representative of the former by retaining a fixed window of beginning and recent tokens in the KV-Cache
but it struggles to flexibly adapt to specific contexts.
E.g., in sentiment analysis, retaining the token ``cute'' in ``a cute cat'' is crucial, while in object recognition, the token ``cat'' would be more important.
H2O~\citep{zhang2024h2o}, on the other hand, employs a token-adaptive approach, using local accumulative attention scores to determine which tokens to evict.
However, it is criticized that in practice, H2O suffers from the attention bias issue~\citep{oren2024transformers}, with a tendency to over-prioritize either the initial or recent tokens.

\begin{figure*}[t]
    \centering
    \includegraphics[width=\linewidth]{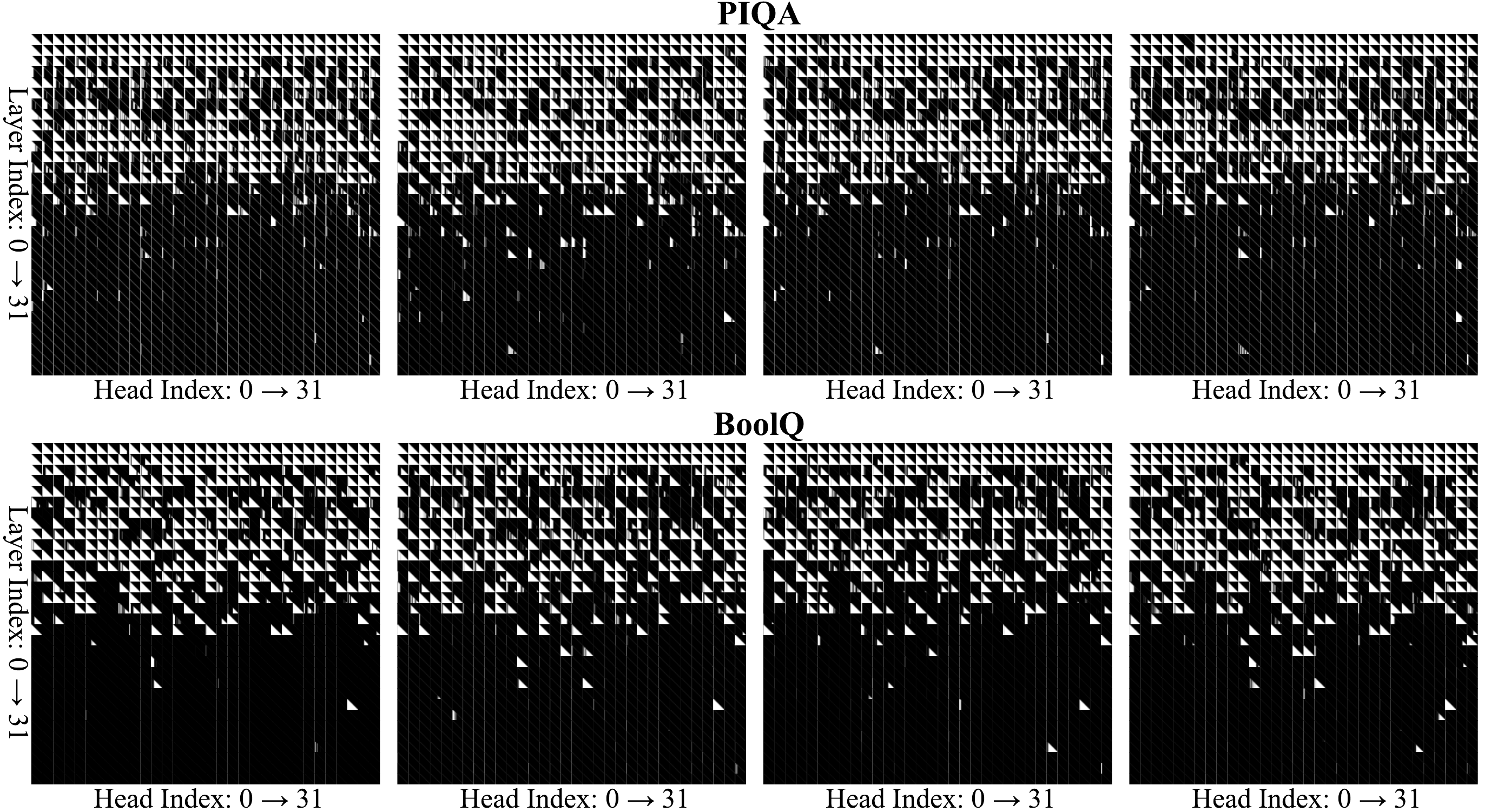}
    \caption{KV-Cache eviction patterns across different layers and attention-heads, visualized for 4 samples from the PIQA dataset (\textbf{top row}) and 4 samples from the BoolQ dataset (\textbf{bottom row}), using AG fine-tuned Llama2-7B models.
    Black areas represent tokens that are neither computed nor stored in the KV-Cache.
    \emph{The variability of eviction patterns across tasks, prompts, layers, and attention-heads demonstrates the dynamic nature of our method}.
    A common trend observed is that deeper layers tend to mask more KV-Cache states, with some in deeper layers being entirely masked.}
    \label{fig:intro}
\end{figure*}

To overcome these challenges, we introduce a parameterized KV-Cache eviction mechanism named Attention-Gate (AG), to perform reliable \emph{in-context} eviction.
AGs are positioned before self-attention layers within the model. They take a sequence of token features as input and generate eviction flags for the tokens, indicating whether the token should be excluded from subsequent self-attention computations.
Tokens that are evicted do not require their KV states to be cached.
AGs can be seamlessly integrated into pre-trained LLMs and tuned by minimizing the language modeling loss.
Ideally, AGs can automatically learn to discern the most relevant tokens for the current context without manual intervention.
In practice, we can implement the AG as a self-attention layer with much fewer heads than the original model (e.g., 4 v.s. 32).
This way, the parallel computational capabilities of the hardware can be harnessed to minimize the extra overhead introduced by AGs.

AG is empirically shown to enjoy high training efficiency, e.g., only four NVIDIA 4090 GPUs and a dataset of 5,000 samples are required for continual pre-training when applying AGs to LLaMA2-7B~\citep{touvron2023llama}.
This alleviates concerns about the computational overhead related to trainable eviction strategies~\citep{zhang2024h2o, chen2024nacl} and amplifies the merits of our approach over existing training-free approaches.
As illustrated in \Cref{fig:intro}, AG generates different eviction strategies across different layers and attention-heads for different tokens, demonstrating its adaptability to the diverse requirements of each component in the model.

To validate the effectiveness of our method, we conduct extensive experiments across multiple benchmarks. After efficient continual pre-training (CPT), our approach outperforms traditional training-free eviction strategies, such as StreamingLLM and H2O, in both accuracy and token eviction rates.
In supervised fine-tuning (SFT), our method not only evicts a significant number of redundant tokens but also maintains or surpasses the performance of LoRA-finetuned LLMs. For example, on the RTE dataset~\citep{bar2006second}, our approach improves accuracy by 13.9\% while evicting 62.8\% of tokens, demonstrating that selective token eviction can enhance performance.
In summary, the Attention-Gate mechanism provides a scalable and efficient solution for KV-Cache management, addressing the limitations of traditional training-free methods.

\section{Related Work}

\label{sec:related}

As large language models (LLMs) scale in both size and input sequence length, optimizing their efficiency has become increasingly important, particularly in addressing space and time complexity. A significant bottleneck lies in the attention mechanism, which demands considerable computational and memory resources, especially for long sequences.

\textbf{Traditional KV-Cache eviction strategies.}
To address both memory and computational challenges, KV-Cache eviction has emerged as an effective strategy.
Existing approaches predominantly rely on parameter-free heuristics.

Static strategies, such as those used in Sparse Transformers~\citep{child2019generating}, employ fixed pruning patterns, such as Strided and Fixed Attention. While effective in some cases, these approaches are not adaptive to specific contexts, often sacrificing accuracy. StreamingLLM~\citep{xiao2024efficientstreaminglanguagemodels} tackles the \emph{Attention Sink} phenomenon, where attention scores concentrate on initial tokens, by retaining these tokens along with a fixed window of recent tokens. While this improves performance, static approaches generally lack the flexibility needed to adapt to different tokens, attention-heads, or layers.

Strategies using accumulative attention scores offer more flexibility by dynamically identifying important tokens. For instance, SpAtten~\citep{wang2021spatten} employs Accumulative Attention Scores (A2S), which sum the softmax outputs for each token to measure its importance. This approach allows selective token pruning in subsequent layers, effectively reducing computational complexity without the need for retraining. H2O~\citep{zhang2024h2o} extends this concept to decoder-based models, using local A2S statistics for adaptive eviction in autoregressive generation. However, H2O suffers from the attention bias issue~\citep{oren2024transformers}, particularly in long-context inputs.
Several follow-up works have aimed to address this limitation. NACL~\citep{chen2024nacl} introduces random eviction to mitigate attention bias, while A2SF~\citep{jo2024a2sf} incorporates a Forgetting Factor. However, none of these approaches fully resolve the underlying problem.
Despite these limitations, some studies~\citep{coldcompress2024} suggest that H2O remains an optimal solution in many scenarios.

\textbf{More adaptive strategies.}
Although strategies based on accumulative attention scores provide more flexibility than static methods, they still have notable limitations. For instance, H2O~\citep{zhang2024h2o} applies the same token eviction ratio across all attention heads, restricting the adaptability of the method.
FastGen~\citep{ge2023model}, on the other hand, introduces a different approach by hybridizing KV-Cache compression policies and applying adaptive strategies to each attention head. However, it focuses on the decoding stage and neglects the importance of the prefilling stage.
Learnable eviction strategies, on the other hand, offer greater flexibility by enabling different layers and attention heads to adopt heterogeneous eviction policies. However, such strategies have been relatively underexplored, likely due to concerns about the computational overhead they may introduce~\citep{zhang2024h2o, chen2024nacl}. Nonetheless, task-specific training is essential for optimizing performance across different contexts. For example, a recent approach~\citep{anagnostidis2024dynamic} introduces a learnable mechanism for dropping uninformative tokens, but it faces difficulties in batched generation and does not account for continual pre-training or decoding-only LLMs.
Despite these challenges, learnable strategies have strong potential to improve performance across a variety of tasks by allowing models to adapt their eviction strategies to meet task-specific requirements.

\section{Method}

\begin{figure}[t]
    \centering
    \includegraphics[width=\linewidth]{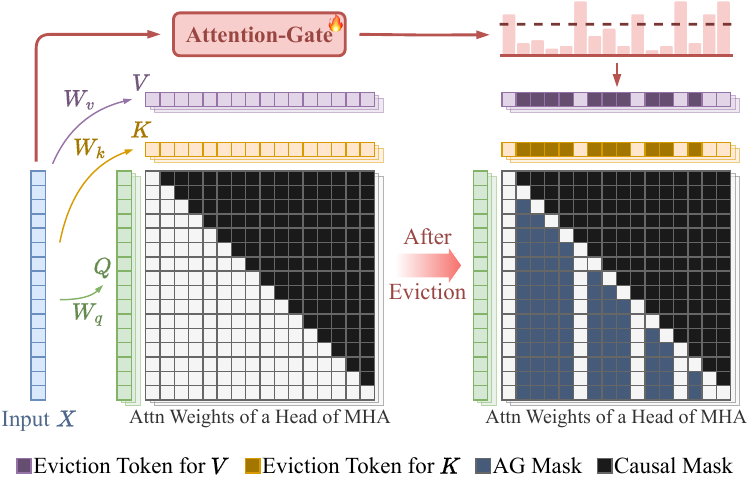}
    \caption{An overview of Attention-Gate (AG) for KV-Cache eviction.
    AG is a lightweight learnable module placed before each MHA layer.
    Given the input hidden states, it determines for each head whether to retain or discard the key and value tokens in the KV-Cache.
    In the attention weights, this corresponds to masking out columns for the evicted keys, while keeping the diagonal intact to ensure the query interacts with its own key.}
    \label{fig:token}

\end{figure}

This section first briefly reveals the basics of multi-head attention and KV-Cache and then describes the proposed \emph{Attention-Gate} (AG) mechanism for in-context KV-Cache eviction for LLM inference acceleration.
An illustrative overview of AG is presented in \Cref{fig:token}.

\subsection{Preliminary}

\textbf{Multi-Head Attention} (MHA)~\citep{vaswani2017attention} is a core component of the Transformer architecture, as used by most LLMs.
MHA enables the model to capture dependencies across different tokens in a sequence.
Specifically, for an input sequence $X \in \mathbb{R}^{n \times d}$, where $n$ represents the sequence length and $d$ denotes the dimensionality of the hidden states, the output of MHA is computed as:
\begin{equation}
    \label{eq:mha}
    \text{MHA}(X) = \left[ \text{H}_1(X), \text{H}_2(X), \dots, \text{H}_h(X) \right] W^O\ ,
\end{equation}
where $\{\text{H}_i\}_{i=1}^{h}$ refers to the $h$ attention heads  and
\begin{align}
    \label{eq:attn-head}
    \text{H}_i(X) &= \text{Attn}\left(XW^Q_i, XW^K_i, XW^V_i\right)\notag\\
    &= \text{Attn}\left(Q_i, K_i, V_i\right)\\
    \label{eq:attn-head}
    &= \text{Softmax}\left( \frac{Q_i K_i^\top}{\sqrt{d_k}} - \text{INF}(\mathbbm{1}-M_i) \right) V_i=A_iV_i.\notag
\end{align}
Here, $W^Q_i, W^K_i \in \mathbb{R}^{d \times d_k}$, $W^V_i \in \mathbb{R}^{d \times d_v}$, and $W^O \in \mathbb{R}^{h d_v \times d}$ are learned projection matrices.
INF is a large constant, $\mathbbm{1}$ is a matrix of ones, $M_i$ is the mask applied to head H$_i$, and $A_i$ represents the attention scores for head H$_i$.

\textbf{KV-Cache} is employed during the inference of auto-regressive transformers, which
stores the key and value information from previous time steps, allowing efficient reuse and reducing recomputation.
The inference process can be divided into two stages: prefilling and decoding.\\
\hlgray{In the prefilling stage}, the input sequence $X^{(\le n)}=\left[x^{(1)},x^{(2)},\dots,x^{(n)}\right] \in \mathbb{R}^{n \times d}$ passes through MHA, and the corresponding key-value pairs $K^{(\leq n)}_i$ and $V^{(\leq n)}_i$ for head H$_i$ are stored in KV-Cache. These are expressed as:
\begin{equation*}
    \label{eq:kv-cache}
    K_i^{(\le n)}=\left[k_i^{(1)},\cdots,k_i^{(n)}\right],\
    V_i^{(\le n)}=\left[v_i^{(1)},\cdots,v_i^{(n)}\right]\ ,
\end{equation*}
where $k_i^{(t)}=x^{(t)}W_i^K$ and $v_i^{(t)}=x^{(t)}W_i^V$.
After prefilling, the next token $x^{(n+1)}$ is generated.\\
\hlgray{In the decoding stage}, $x^{(n+1)}$ is input to generate $x^{(n+2)}$ for the first step.
During this process, only $k_i^{(n+1)},v_i^{(n+1)}$ need to be computed for head H$_i$.
These are then concatenated with the cached $K_i^{(\leq n)}$ and $V_i^{(\leq n)}$ to form $K_i^{(\leq n+1)}$ and $V_i^{(\leq n+1)}$, which are used to complete the current MHA computation and update the KV-Cache.
The process repeats token by token until the sequence generation is complete.

KV-Cache plays a critical role in improving the efficiency of LLM inference.
However, the size of the KV-Cache grows with the input sequence length, leading to substantial memory overhead. Efficiently managing KV-Cache while maintaining model performance has become a key challenge in scaling LLMs to longer contexts.

\subsection{Limitations of Traditional Eviction Strategies}

\label{sec:lim}

\textbf{Lack of flexibility.}
Static KV-Cache eviction strategies, such as those used in StreamingLLM~\citep{xiao2024efficientstreaminglanguagemodels}, lack adaptability across key dimensions, including token-specific, attention-head-specific, layer-specific, task-specific, and model-specific contexts.
H2O~\citep{zhang2024h2o} addresses some of these limitations by introducing token-level and head-level adaptability. However, it still applies a uniform eviction ratio across all attention heads, which limits its granularity at the head level.
This isotropic approach overlooks the significant variation in attention scores across different heads, as demonstrated by FastGen~\citep{ge2023model}.
This observation underscores the need for eviction strategies that are adaptive across multiple dimensions to better handle the diverse contexts processed by the model.
Without finer-grained flexibility, existing training-free strategies risk retaining unnecessary information, leading to higher memory usage and reduced efficiency.

\textbf{Absence of global statistics.}
KV-Cache eviction strategies should ideally be context-aware, as the importance of the same token can vary significantly depending on the surrounding context.
To accurately discard redundant or less important tokens, it is necessary to consider the global statistics of the context. Strategies relying on accumulative attention scores, such as H2O, NACL, and A2SF~\citep{chen2024nacl, jo2024a2sf, zhang2024h2o}, are primarily based on local statistics of the context.
While methods like NACL and A2SF attempt to mitigate the limitations of local context by introducing various adjustments to reduce the misjudgment of token importance and the resulting biased retention~\citep{oren2024transformers}, they fail to address the root issue.
The core problem lies in the lack of consideration of the \emph{global} context when determining which tokens to evict.

\textbf{Inefficiency.}
Methods like H2O and FastGen~\citep{ge2023model} are inefficient due to their sequential, token-by-token eviction processes at each decoding step. H2O, for example, computes attention scores before deciding which tokens to evict, wasting computation on soon-to-be discarded tokens.

Our method tries
to address the aforementioned limitations.

\subsection{Attention-Gate}

\label{sec:ag}

The \emph{Attention-Gate} (AG) is a lightweight, trainable module positioned before the MHA layer, generating \emph{eviction flags} for the tokens.
The flags determine which tokens in the KV-Cache of each head of the MHA should join the computation of attention scores and be retained in the KV-Cache.  %

\textbf{Main architecture.}
AG consists of \hlblue{(i)} an attention-like structure and \hlpurple{(ii)} a gating mechanism.
\hlblue{(i)} A modified version of the standard MHA in \Cref{eq:mha}, denoted as $\text{MHA}'$, is used to facilitate \emph{global information} exchange across all tokens in the sequence.
To distinguish it from the vanilla MHA, all symbols in $\text{MHA}'$ are marked with a prime ($\prime$).
Notably, $\text{MHA}'$ has much fewer heads compared to MHA, i.e., $h' < h$, and its projection matrix $W^{O'} \in \mathbb{R}^{h' d'_k \times h}$ differs slightly from $W^O \in \mathbb{R}^{h d_v \times d}$.
\hlpurple{(ii)} The gating mechanism, denoted as G, is then introduced to enable adaptive eviction policies for attention heads within the MHA layer.

\textbf{Input \& output.}
AG takes the hidden states as input and generates binary flags as output, one for each token in every attention head.
Specifically, given an input sequence $X\in\mathbb{R}^{n\times d}$, the output of AG is computed as:
\begin{equation}
    \label{eq:attention-gate}
    \text{AG}(X) = \text{G}\left(\text{MHA}'(X), \tau\right)\ \in\{0,1\}^{n\times h},
\end{equation}
where \(\tau\) is a hyperparameter.
The gate G outputs 1 or 0 based on the score $s$ of a token obtained from MHA$'$:
\begin{equation}
    \label{eq:gate}
    \text{G}(s,\tau)=\begin{cases}
        1,&\text{if }\text{Sigmoid}(s)> \tau\\
        0,&\text{otherwise}
    \end{cases}\ .
\end{equation}
The values $k_i^{(t)}$ and $v_i^{(t)}$ for token $x^{(t)}$ and attention-head H$_i$ are retained if $\text{AG}(X)_i^{(t)} = 1$; otherwise, they are discarded.
Based on this, the mask $M_i = \big[m_i^{(j,t)}\big]$ for the attention scores $A_i$ of head H$_i$ in \Cref{eq:attn-head} is defined as:
\begin{equation}
    \label{eq:attn-scores}
    m_i^{(j,t)}=\begin{cases}
        0,&\text{if } j < t\\
        1,&\text{if } j = t\\
        \text{AG}(X)_i^{(t)}\ \ ,&\text{otherwise}
    \end{cases}\ .
\end{equation}
In the attention matrix, the columns corresponding to evicted tokens are masked out. However, \emph{the diagonal elements, where a token attends to itself, are always preserved}.
This ensures that each token maintains interaction with its own key, guaranteeing self-attention remains intact.

In this way, the AG module selectively determines which tokens are retained or discarded for each attention head, based on the global information captured by $\text{MHA}'$ and the gating mechanism.

\subsection{Design Rationale}
\label{sec:design}

\textbf{Local or global?}
Our AG design, introduced in \Cref{sec:ag}, utilizes a multi-head attention mechanism (i.e. MHA$'$) that inherently incorporates global information.
This approach allows the model to aggregate context across the entire sequence, enabling eviction decisions to accurately reflect the broader in-context information. Such a design is particularly effective for assessing the importance of tokens, as determining redundancy or relevance often requires a global understanding of the sequence.

Alternatively, a simpler approach can be used, such as employing a linear transformation to generate eviction flags.
This method relies solely on local information, where each token only considers its own hidden state without incorporating information from other tokens in the sequence.
While the local approach has the advantage of being computationally cheaper, as shown in \Cref{tab:comparison}, its performance and efficiency are not guaranteed. This limitation highlights the challenges faced by methods rely on local statistics of the context, as discussed in \Cref{sec:lim}.

\vspace{10pt}
\textbf{Softmax or Sigmoid?}
In our design, as described in \Cref{eq:gate}, we use the Sigmoid function to calculate the eviction probability for each token.
However, it is worth noting that Softmax is another common choice for generating probabilities, as seen in applications like the gating mechanisms of Mixture of Experts in LLMs.

While both functions are viable, they offer different trade-offs. The Sigmoid function, as used in our design, computes probabilities \emph{independently} for each token. This independence is important because it ensures that a token’s eviction likelihood is not directly influenced by the probabilities of other tokens in the sequence, allowing for a more flexible and precise eviction policy.

On the other hand, Softmax normalizes the probabilities across the entire sequence, which introduces competition between tokens. This can be beneficial in scenarios where the relative importance of tokens needs to be considered. However, in the context of eviction, where we want to assess each token individually without direct dependencies on others, Softmax might not be the ideal choice.

\vspace{10pt}

\textbf{Prefilling or decoding?}
AG is primarily applied during the prefilling stage, where the full sequence is available.
By making eviction decisions before the MHA layers, AG effectively manages the KV-Cache during this phase.
In contrast, AG is not used during the decoding stage to avoid adding complexity to the inference process.
Using AG in decoding could slow down inference and increase the cache memory footprint, as additional keys and values from the attention-like structure would need to be stored.

\subsection{Training Implementation}
\label{sec:training}

To train the Attention-Gate (AG) effectively, this section outlines the key components of the training process. For more details, please refer to \Cref{sec:exp} and \Cref{sec:appendix}.

\textbf{Eviction Loss.}
To encourage the eviction of unnecessary tokens, we introduce a dedicated loss function called the \emph{Eviction Loss}.
The loss function encourages the model to maintain the eviction ratio close to a target value
\(\beta\),
which is defined as:
\begin{equation}
\label{eq:evict}
\ell_{\text{evict}} = \alpha \cdot \left|\overline{\text{AG}} - \beta \right|,
\end{equation}
where
$\overline{\text{AG}}$ represents the average output of all AG modules.
In this formula, $\alpha$ adjusts the intensity of KV-Cache eviction, while $\beta \in [0, \tau]$ ensures that eviction does not become overly aggressive.
Eviction Loss allows for adaptive eviction across layers and attention-heads.
This loss function works alongside the auto-regressive loss to balance token eviction with maintaining model performance.

\textbf{Initialization.}
We initialize the AG parameters using Xavier initialization~\citep{glorot2010understanding} to provide a stable starting point for learning.
Additionally, a small constant $\gamma \geq 0$ can optionally be added inside Sigmoid in \Cref{eq:gate}, ensuring that the initial retention probabilities are close to 1.
This encourages the model to retain most tokens early in training.

\textbf{Handling non-differentiability.}
Directly applying the threshold-based gating mechanism from \Cref{sec:ag} would lead to non-differentiable gradients during training due to the hard thresholding’s discrete nature.
To resolve this, we employ the \emph{Straight-Through Estimator} (STE)~\citep{yin2019understanding}, which allows gradients to flow through discrete decisions by approximating them during the backward pass.
Specifically, during backpropagation, instead of using the hard 0 or 1 values obtained from comparing against the threshold, we utilize the smooth output of the Sigmoid function.
This approach ensures smooth gradients and enables effective training of the AG while preserving its binary behavior during the forward pass.

\begin{table*}[t]
    \caption{Performance comparison of Llama2-7B and various KV-Cache eviction strategies \emph{after continual pre-training}.
    For baselines, ($W_q, W_k, W_v, W_o$) are made trainable, while in our method, the AG module is also trainable.
    Higher values indicate better performance for all metrics.
    Acc. refers to accuracy.
    \%Evict. refers to the mean KV-Cache eviction ratio, representing the percentage of tokens evicted from KV-Cache.
    The eviction ratio is fixed at 50\% for the baseline methods.
    In contrast, our method
    achieves better performance (average accuracy and score) while maintaining a higher average \%Evict..
    }
    \label{tab:app_cpt}
    \setlength{\tabcolsep}{3.3pt} %
    \small
    \centering
    \begin{tabular}{ccccccccccc|cc} %
        \toprule
        ~ & Metric & PIQA & ARC-C & ARC-E & RTE & COPA & BoolQ & HellaSwag & MMLU & Avg. & Metric & LongBench \\
        \midrule
        Llama2-7B-cpt & Acc. & 72.69 & 32.88 & 50.62 & 50.54 & 57.00 & 64.77 & 42.19 & 26.64 & 49.67 & Score & 23.42 \\
        \midrule
        StreamingLLM & Acc. & 72.42 & 31.53 & \textbf{49.74} & 50.90 & 54.00 & 61.31 & 37.75 & 26.66 & 48.04 & Score & 4.61 \\
        H2O & Acc. & 72.20 & 30.85 & 49.38 & \textbf{51.99} & 55.00 & \textbf{62.42} & 41.45 & 26.45 & 48.72 & Score & 4.85 \\
        \midrule
        \multirow{2}{*}{Ours} & Acc. & \textbf{76.33} & \textbf{32.20} & 48.32 & 50.18 & \textbf{59.00} & 60.46 & \textbf{64.23} & \textbf{28.54} & \textbf{52.41} & Score & \textbf{13.71} \\
        ~ & \%Evict. & 43.12 & 46.54 & 45.15 & 48.60 & \textbf{55.37} & \textbf{50.16} & \textbf{61.10} & \textbf{70.36} & \textbf{52.55} & \%Evict. & \textbf{68.55} \\
        \bottomrule
    \end{tabular}
\end{table*}

\subsection{Complexity Analysis}
\label{sec:comp}

\textbf{AG module.}
For input \( X \in \mathbb{R}^{n \times d} \), the FLOPs for AG are:
\[
\operatornamewithlimits{\text{FLOPs}}_{\text{AG}} = \mathcal{O}(n^2d_k'h')\ .
\]
\textbf{MHA module.}
Without AG, original MHA's FLOPs are:
\begin{equation}
\label{eq:flopswoag}
\operatornamewithlimits{\text{FLOPs}}_{\text{original MHA}} = \mathcal{O}(n^2d_kh)\ .
\end{equation}
After AG processing, \( t\% \) of the KV-Cache tokens are discarded, leaving \( (1 - t\%) \) for attention computation:
\[
\operatornamewithlimits{\text{FLOPs}}_{\text{MHA after AG}} = \mathcal{O}\big((1-t\%)n^2d_kh\big)\ .
\]
\textbf{Combined AG \& MHA}, the total FLOPs are:
\begin{equation}
    \label{eq:flops}
    \operatornamewithlimits{\text{FLOPs}}_{\text{AG \& MHA}} = \mathcal{O}\big(n^2\max\big(d_k'h', (1-t\%)d_kh\big)\big)\ .
\end{equation}
\textbf{Efficiency.}
The reduction in FLOPs depends on three factors:
\hlpurple{(i)} Reduction in token count ($ t\% $): Higher values of \( t\% \) result in a larger reduction in the quadratic term of the original MHA.
\hlpurple{(ii)} Head configuration ($ h' < h $): The AG module must have significantly fewer heads (\( h' \)) compared to the original MHA (\( h \)) to ensure its overhead is small.
\hlpurple{(iii)} Head dimension ratio ($ d_k' < d_k $): A smaller head dimension (\( d_k' \)) for AG further reduces its contribution to total FLOPs.

\section{Experiments}

\label{sec:exp}

This section consists of three main parts.
First, we evaluate the performance of AG in two scenarios: continual pre-training (CPT) and supervised fine-tuning (SFT) (\Cref{sec:cpt} \& \ref{sec:sft}).
Second, we provide visualization of selected examples to demonstrate the core characteristics of AG (\Cref{sec:visual}).
Finally, we conduct ablation studies to provide further insights into the effectiveness of AG (\Cref{sec:ablation}).
Additional results are provided in \Cref{sec:app_cpt}.

\subsection{Continual Pre-training}

\label{sec:cpt}

\subsubsection{Setup}

\label{sec:cpt-setup}

\textbf{Models \& datasets.}
We use Llama2-7B~\citep{touvron2023llama} as our primary base model.
Additionally, we validate the feasibility of our approach on Mistral-7B~\citep{jiang2023mistral}, with results provided in \Cref{tab:mistral}.
For continual pre-training, we select a subset of the RedPajama dataset~\citep{together2023redpajama}, comprising approximately 5,000 samples~\footnote{Specifically, we sampled 4,997 samples proportionally from each subset of the RedPajama dataset.}, to serve as the training set.
To assess the effectiveness of our method, we evaluate it on widely recognized benchmarks: PIQA~\citep{Bisk2020}, ARC-C~\citep{allenai:arc}, ARC-E~\citep{allenai:arc}, RTE~\citep{bar2006second}, COPA~\citep{roemmele2011choice}, BoolQ~\citep{clark2019boolq}, HellaSwag~\citep{zellers2019hellaswag}, MMLU~\citep{hendryckstest2021} and LongBench~\citep{bai2023longbench}.
All evaluations are conducted in a zero-shot setting, with performance assessed using OpenCompass~\citep{2023opencompass}.

\textbf{Training details \& baselines.}
For the Llama2-7B model, the following hyperparameters from \Cref{sec:training} were used: $\tau = 0.5$, $\gamma = 2$, $\alpha = 5$, and $\beta = 0.4$. The model was trained for a single epoch.
The evaluation metrics include performance (accuracy and score) and the mean eviction ratio for all KV-Cache. For KV-Cache eviction strategies, we use StreamingLLM~\citep{xiao2024efficientstreaminglanguagemodels} as a representative of static strategies, while H2O~\citep{zhang2024h2o} represents methods based on accumulative attention scores~\footnote{\citet{coldcompress2024} suggest that H2O is an optimal solution in many scenarios. Due to its strong performance and the difficulty of reproducing other training-free methods, no additional training-free baselines were included.}.
For baseline methods, eviction ratio is fixed at 50\%, and ($W_q, W_k, W_v, W_o$) are made trainable using LoRA~\citep{hu2021lora}. In our method, AG is also trainable~\footnote{We tested a version where only AG is trainable, with other parameters frozen. Results are shown in \Cref{tab:cpt}.}.

\subsubsection{Results}

The results in \Cref{tab:app_cpt} demonstrate the effectiveness of our method in balancing performance and KV-Cache eviction.
Our method consistently outperforms the baseline strategies in terms of performance across most tasks while achieving a higher mean eviction ratio.

Moreover, it is worth highlighting that our method achieves these results with minimal computational overhead, as detailed in \Cref{sec:cpt-setup}.
The continual pre-training was conducted on only 5,000 samples and trained for just one epoch, demonstrating the lightweight nature of our approach.
This efficiency can be attributed to the fact that our method does not need to learn new knowledge from scratch but rather focuses on learning effective token retention strategies, leveraging the existing capabilities of the pre-trained model.

\begin{table}[t]
    \caption{Performance of Llama2-7B with LoRA fine-tuning and our method on six downstream tasks.
    In addition to the LoRA fine-tuning targets, our method makes the AG modules learnable.
    Two settings for $\alpha$ (0.5 and 1) are tested.
    Our method maintains comparable or better accuracy while achieving a higher eviction ratio, demonstrating its \emph{task-specific adaptability} in managing token eviction.
    }
\label{tab:sft}
\setlength{\tabcolsep}{1.3pt}
\small
\centering
\resizebox{0.45\textwidth}{!}{
  \begin{tabular}{ccccccccc}
    \toprule
    ~ & Metric & PIQA & ARC-C & RTE & COPA & BoolQ & OBQA & Avg.\\
    \midrule
    {\scriptsize Fine-tuned} & \multirow{2}{*}{Acc.} & \multirow{2}{*}{82.92} & \multirow{2}{*}{60.34} & \multirow{2}{*}{64.98} & \multirow{2}{*}{92.00} & \multirow{2}{*}{88.10} & \multirow{2}{*}{78.80} & \multirow{2}{*}{77.86} \\
    {\scriptsize Llama2-7B} & ~ & ~ & ~ & ~ & ~ & ~ & ~ & ~ \\
    \midrule
    Ours & Acc. & 82.15 & 59.66 & 64.26 & 93.00 & 86.82 & 78.80 & 77.45 \\
    {\scriptsize $(\alpha=1)$} & \%Evict. & 66.16 & 48.31 & 65.47 & 45.40 & 67.46 & 67.17 & 60.00 \\
    \midrule
    Ours & Acc. & 81.50 & 57.63 & 74.01 & 95.00 & 87.00 & 79.20 & 79.06 \\
    {\scriptsize $(\alpha=0.5)$} & \%Evict. & 64.96 & 36.45 & 62.80 & 34.77 & 67.31 & 66.65 & 55.49 \\
    \bottomrule
  \end{tabular}
}
\vspace{-10pt}
\end{table}

\subsection{Supervised Fine-tuning}

\label{sec:sft}

\subsubsection{Setup}

\label{sec:sft-setup}

\textbf{Model \& tasks.}
We utilize Llama2-7B~\citep{touvron2023llama} as base model.
To evaluate our approach, we select six widely recognized downstream tasks: PIQA~\citep{Bisk2020}, ARC-C~\citep{allenai:arc}, RTE~\citep{bar2006second}, COPA~\citep{roemmele2011choice}, BoolQ~\citep{clark2019boolq}, and OpenBookQA~\citep{OpenBookQA2018}.
For each task, we fine-tune model using the respective training set and evaluate its performance on the corresponding test set.

\textbf{Implementation details.}
The selection of trainable parameters follows \Cref{sec:cpt-setup}.
The hyperparameters are $\tau = 0.3$, $\gamma = 0$, $\alpha = 1$ or $0.5$, and $\beta = 0.28$. Training is performed using the AdamW optimizer~\citep{loshchilov2017decoupled} with a learning rate of 5e-5 for 2 epochs per dataset.

\subsubsection{Results}

As shown in \Cref{tab:sft}, our method achieves a strong balance between accuracy and KV-Cache eviction.

With $\alpha=1$, it maintains competitive accuracy compared to the fine-tuned Llama2-7B baseline
while achieving a high mean eviction ratio of 60.00\%. With $\alpha=0.5$, the eviction ratio decreases to 55.49\%, but the average accuracy improves
. In tasks like RTE and COPA, it even surpasses the baseline. This suggests that effective token eviction helps the model focus on relevant information.

Additionally, performance varies across tasks under the same settings. For instance, ARC-C is more challenging to evict compared to OpenBookQA, leading to a larger accuracy drop post-eviction. This highlights the importance of \emph{task-specific} KV-Cache eviction policies.

\begin{figure}[t]
    \centering
    \includegraphics[width=0.9\linewidth]{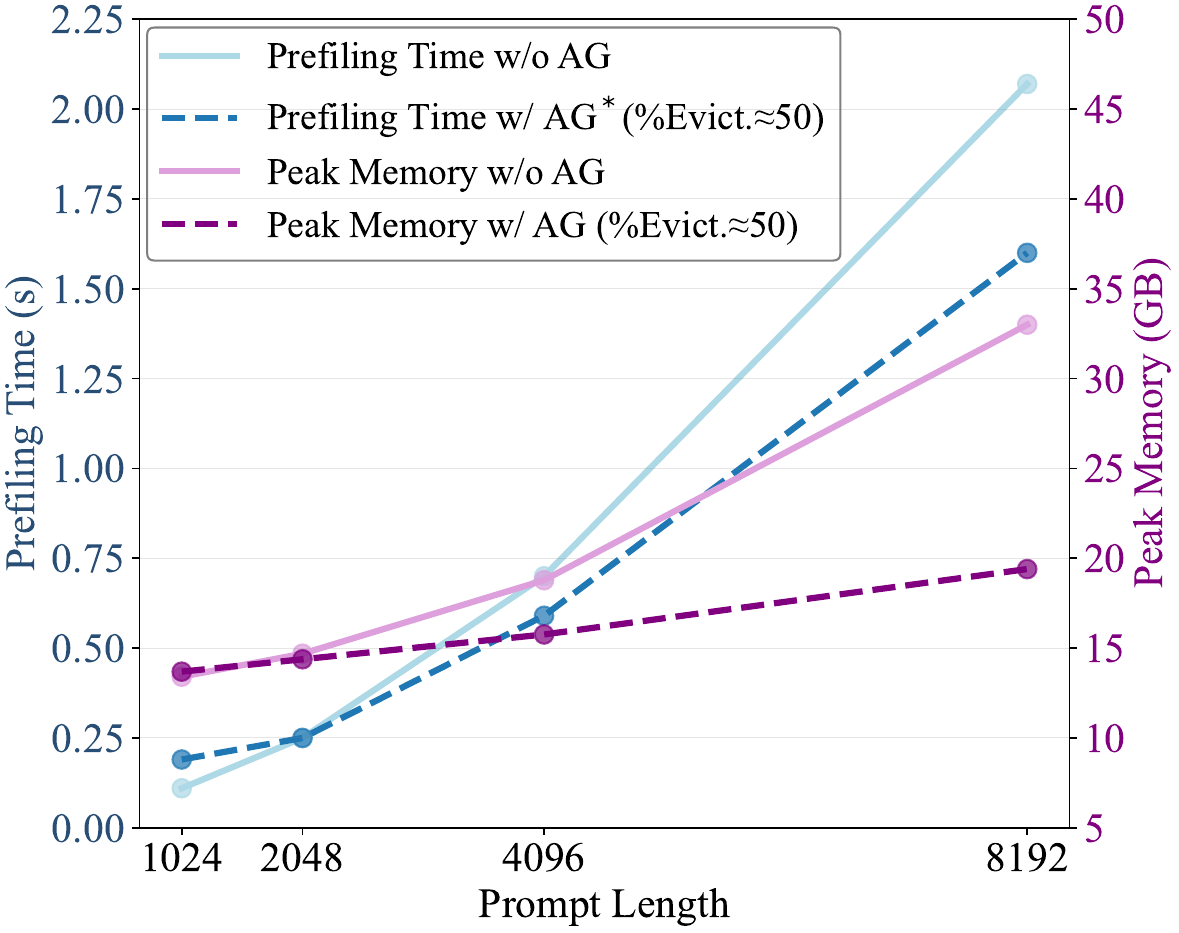}
    \caption{
Comparison of peak memory usage and prefilling time between the LLaMA2-7B model (without AG) and the proposed implementation (with AG and $\sim$50\% eviction) across varying prompt lengths. The results show significant improvements in memory efficiency with AG, especially as prompt length increases. Prefilling time is not the primary focus, and the current implementation (marked with * in the legend) relies on a suboptimal for-loop over attention heads. Even so, the method maintains stable prefilling time and shows a clear reduction trend with longer prompts.
}
    \label{fig:complexity}
    \vspace{-10pt}
\end{figure}

\begin{figure}[t]

    \centering
    \includegraphics[width=0.95\linewidth]{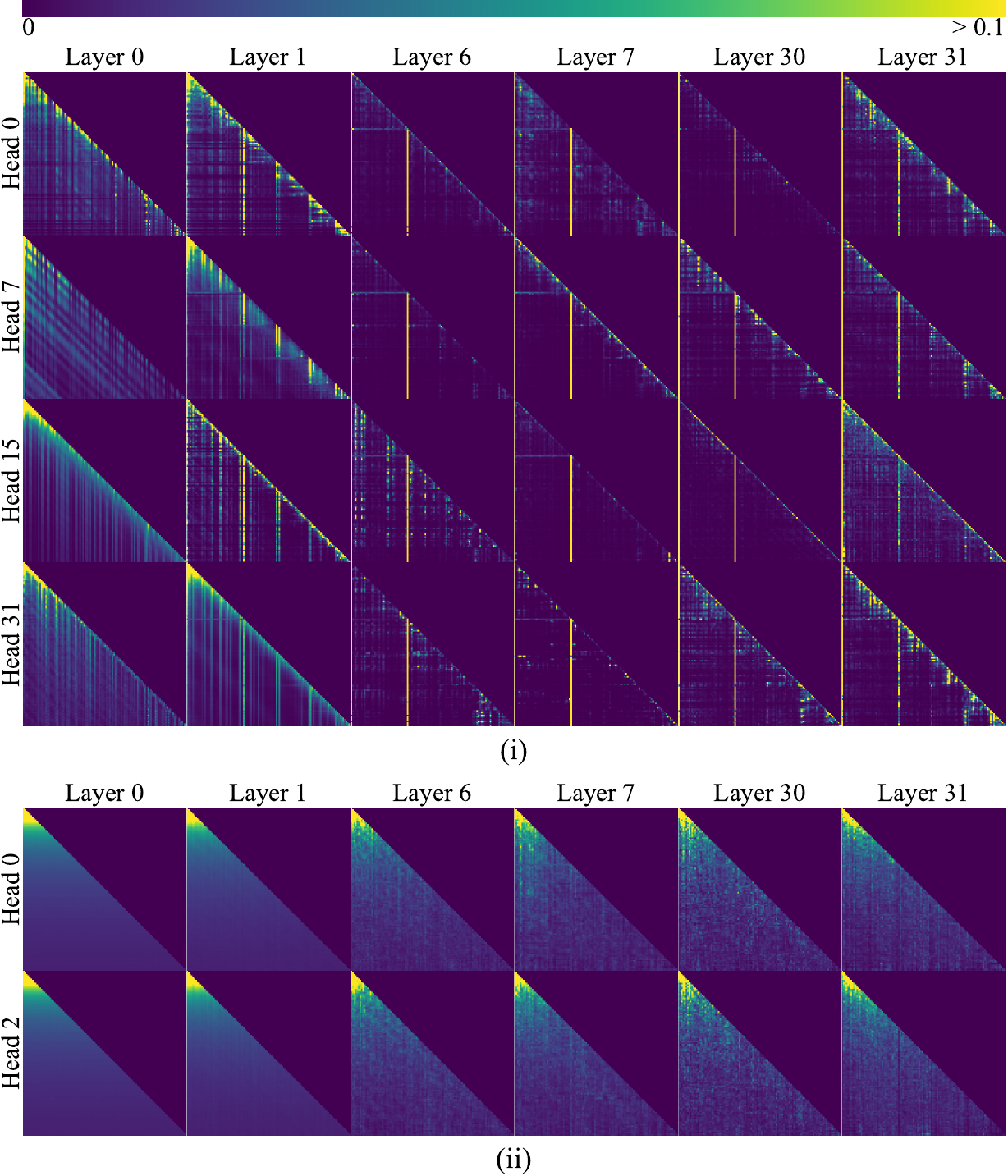}
    \caption{This visualization highlights attention patterns in Llama2-7B after fine-tuning on the BoolQ dataset, showcasing multiple heads within both MHA and AG across different layers using a selected sample.
    \hlgray{In part (i)}, we visualize attention scores from several MHA heads across layers before eviction:
    \hlpurple{1.} MHA heads display diverse attention patterns, especially in the first two layers, where heterogeneity is prominent.
    \hlpurple{2.} Attention patterns become progressively sparser in deeper layers, transitioning from dense in early layers.
    \hlpurple{3.} Bright-yellow vertical lines, indicating critical tokens for inference, appear consistently across heads in deeper layers. These align with the \emph{Heavy Hitters} in H2O~\citep{zhang2024h2o}, highlighting tokens that significantly contribute to attention scores. Our method ensures these critical tokens are preserved in deeper layers, maintaining their importance across the network.
    \hlgray{In part (ii)}, we visualize the attention-like scores from the AG mechanism:
    As layers deepen, AG shifts from high-resolution to lower-resolution attention, focusing on distilling in-context information. Deeper AG layers no longer require high resolution for capturing global context, as earlier layers have already refined it. This suggests potential optimizations, such as reducing the number of heads or dimensions in deeper AG layers, to enhance efficiency.
    }
    \label{fig:visual}
    \vspace{-20pt}
\end{figure}

\subsubsection{Inference Efficiency}

We evaluated the inference efficiency of the LLaMA2-7B model with and without the AG module. As shown in \Cref{fig:complexity}, our method achieves notable improvements in memory efficiency, especially with longer prompts. Although prefilling time is influenced by the current implementation, it remains stable and demonstrates scalability as prompt lengths increase.
We believe there is room for acceleration using kernel fusion
techniques and leave it for future work.

\subsection{Visualization}

\label{sec:visual}

\begin{table}[t]
    \caption{Ablation study on various settings of AG, reporting accuracy (Acc.) and KV-Cache eviction ratio (\%Evict.) under different configurations, with $\alpha=1$ in all settings.
    \hlgray{The results for (1)} correspond to the setup described in \Cref{sec:sft}, where the number of AG heads is 4, and the head dimension is 128.
    \hlgray{In (2-1) and (2-2)}, we explore the impact of the number of AG heads, with (2-1) using 2 heads and (2-2) using 1 head.
    The comparison between (1), (2-1), and (2-2) shows that reducing the number of heads leads to a drop in both accuracy and eviction ratio, indicating that the capacity of AG is closely tied to the number of heads.
    \hlgray{For (3-1) and (3-2)}, we assess the effect of reducing head dimensions for AG heads, where (3-1) has half the dimension size of (1) and (3-2) has 1/4.
    Comparing (1), (3-1), and (3-2) reveals that smaller dimensions reduce the eviction capability and accuracy, highlighting the importance of maintaining sufficient head dimensionality.
    \hlgray{Settings (4-1) and (4-2)} explore using the previous layer’s hidden states and AG module to guide the current layer’s eviction strategy. In (4-1), eviction begins from the second layer onward, while in (4-2), eviction starts from the third layer, reflecting the denser attention patterns in the first two layers (\Cref{sec:visual}). Both settings show a slight decline in accuracy and eviction ratio compared to (1) but introduce parallelism, suggesting potential for future optimization.
    \hlgray{The setting (5)} replaces MHA$'$ in AG with a simple linear layer to determine the eviction strategy.
    The comparison with (1) shows that linear layers almost cannot evict tokens effectively, reinforcing the necessity of leveraging global in-context information for successful eviction, as discussed in \Cref{sec:design}.}
    \label{tab:comparison}
    \centering
\resizebox{0.47\textwidth}{!}{
\setlength{\tabcolsep}{4pt}
    \begin{tabular}{cccccccccc}
        \toprule
        ~ & Metric & PIQA & ARC-C & RTE & COPA & BoolQ & OBQA & Avg. \\
        \midrule
        \multirow{2}{*}{(1)} & Acc. & 82.15 & 59.66 & 64.26 & 93.00 & 86.82 & 78.80 & 77.45 \\
        ~ & \%Evict. & 66.16 & 48.31 & 65.47 & 45.40 & 67.46 & 67.17 & 60.00 \\
        \midrule
        \multirow{2}{*}{(2-1)} & Acc. & 81.88 & 57.63 & 65.70 & 91.00 & 87.52 & 77.40 & 76.86 \\
        ~ & \%Evict. & 63.92 & 36.38 & 62.73 & 24.38 & 65.22 & 63.57 & 52.70 \\
        \multirow{2}{*}{(2-2)} & Acc. & 82.15 & 53.90 & 62.45 & 89.00 & 87.31 & 77.40 & 75.37 \\
        ~ & \%Evict. & 58.97 & 31.47 & 59.77 & 20.32 & 63.02 & 59.17 & 48.79 \\
        \midrule
        \multirow{2}{*}{(3-1)} & Acc. & 81.45 & 53.36 & 58.84 & 88.00 & 86.73 & 78.40 & 74.46 \\
        ~ & \%Evict. & 61.75 & 33.55 & 61.34 & 19.24 & 64.59 & 59.59 & 50.01 \\
        \multirow{2}{*}{(3-2)} & Acc. & 83.03 & 53.90 & 59.93 & 89.00 & 87.16 & 76.40 & 74.90 \\
        ~ & \%Evict. & 58.68 & 24.23 & 32.23 & 12.28 & 59.40 & 55.54 & 40.39 \\
        \midrule
        \multirow{2}{*}{(4-1)} & Acc. & 81.66 & 55.25 & 66.06 & 88.00 & 86.85 & 78.00 & 75.97 \\
        ~ & \%Evict. & 49.52 & 36.92 & 46.85 & 28.74 & 56.02 & 60.32 & 46.40 \\
        \multirow{2}{*}{(4-2)} & Acc. & 82.75 & 55.93 & 79.06 & 82.00 & 86.33 & 78.40 & 77.41 \\
        ~ & \%Evict. & 53.31 & 44.38 & 51.20 & 47.95 & 61.98 & 61.73 & 53.43 \\
        \midrule
        \multirow{2}{*}{(5)} & Acc. & 82.54 & 54.58 & 57.40 & 81.00 & 87.71 & 74.80 & 73.01 \\
        ~ & \%Evict. & 1.06 & 0.46 & 0.81 & 0.26 & 1.38 & 1.16 & 0.86 \\
        \bottomrule
    \end{tabular}
    }
    \vspace{-20pt}
\end{table}

In this section, we present visualizations to highlight key characteristics of our AG mechanism.
After fine-tuning on specific tasks, we visualize the model’s MHA and AG weights for selected samples, as shown in \Cref{fig:visual}.
The complete version can be found in \Cref{sec:more_visual}.

\subsection{Ablation}
\label{sec:ablation}

In the ablation study, we investigate the effects of various configurations on the AG mechanism, including the number of AG heads, head dimensions, and eviction strategies. Detailed settings and performance results are provided in \Cref{tab:comparison}.
Key findings include the impact of reducing the number of heads (as seen in (2-1) and (2-2)) and head dimensions (in (3-1) and (3-2)), both of which lead to lower accuracy and eviction ratios. In (4-1) and (4-2), we evaluate eviction strategies where the current layer’s eviction is based on the previous layer’s hidden states and AG module, introducing parallelism but with a slight performance trade-off. Replacing MHA$'$ with a linear layer in setting (5) highlights the importance of an attention-like structure for effective token eviction.
Additional results are provided in \Cref{tab:comparison2}.

\section{Conclusion}

In conclusion, the proposed Attention-Gate mechanism offers a flexible and adaptive solution to KV-Cache eviction in large language models.
By dynamically identifying and discarding less important tokens in a data-driven manner, Attention-Gate addresses the limitations of static and attention-score-based strategies, providing efficient context-aware eviction.
This mechanism integrates seamlessly with pre-trained models and can be easily tuned, making it a practical and effective method for enhancing both performance and memory efficiency in various tasks.

\section*{Impact Statement}

This paper presents work whose goal is to advance the field of Machine Learning. There are many potential societal consequences  of our work, none which we feel must be specifically highlighted here.

\bibliography{main}
\bibliographystyle{icml2025}

\newpage
\appendix
\onecolumn

\section{Additional Experiments}
\label{sec:appendix}

{
\subsection{Additional Results for Continual Pre-training}
\label{sec:app_cpt}

In this section, we perform continual pre-training on Llama2-7B using the same training data and hyperparameter settings described in \Cref{sec:cpt-setup}. The baselines are training-free, while in our method, only the AG module is trainable. The results are shown in \Cref{tab:cpt}.

\begin{table*}[h]
    \caption{Performance comparison of Llama2-7B and various KV-Cache eviction strategies across seven tasks.
    \emph{Our approach trains only the AG module during continual pre-training, keeping other components frozen.}
    The table reports accuracy (Acc.) for Llama2-7B and all eviction methods, with Llama2-7B serving as the upper bound for accuracy.
    Metric \%Evict. refers to the mean KV-Cache eviction ratio, representing the percentage of tokens evicted from the KV-Cache.
    The eviction ratio is fixed at 50\% for the baseline methods, including a local strategy (retaining only recent tokens), StreamingLLM, and H2O.
    In contrast, our method achieves higher average accuracy while maintaining a higher average \%Evict..
    }
    \label{tab:cpt}
    \small
    \centering
    \begin{tabular}{cccccccccc} %
        \toprule
        ~ & Metric & PIQA & ARC-C & ARC-E & RTE & COPA & BoolQ & HellaSwag & Avg. \\
        \midrule
        Llama2-7B & Acc. & 76.33 & 37.29 & 51.32 & 51.99 & 62.00 & 69.94 & 68.16 & 59.58 \\
        \midrule
        Local & Acc. & 69.97 & 31.86 & 48.68 & 51.99 & 60.00 & 57.86 & 37.08 & 51.06 \\
        StreamingLLM & Acc. & 72.69 & 33.22 & 51.15 & 50.18 & \textbf{63.00} & 62.05 & 40.85 & 53.31 \\
        H2O & Acc. & 75.9 & 33.22 & \textbf{52.03} & \textbf{52.71} & 47.00 & 67.37 & 66.32 & 56.36 \\
        \midrule
        \multirow{2}{*}{Ours} & Acc. & \textbf{76.17} & \textbf{33.90} & 49.03 & 52.35 & \textbf{63.00} & \textbf{67.52} & \textbf{66.33} & \textbf{58.33} \\
        ~ & \%Evict. & \textbf{54.29} & \textbf{51.03} & \textbf{51.05} & 46.70 & 40.02 & \textbf{57.75} & \textbf{52.16} & \textbf{51.87} \\
        \bottomrule
    \end{tabular}
\end{table*}

\subsection{Results of Continual Pre-training on Mistral}

We conducted continual pre-training on Mistral-7B~\citep{jiang2023mistral} using 5,000 samples from RedPajama~\citep{together2023redpajama}, and the results are shown in \Cref{tab:mistral}. Compared to the performance of Llama2-7B presented in \Cref{tab:cpt}, Mistral's performance slightly declined. We hypothesize that this may be due to the distribution of RedPajama's data being less suited to Mistral. Additionally, this raises the question of whether KV-Cache eviction is model-dependent, and whether its effectiveness is related to the model's expressive power.
Although the parameter counts of Mistral-7B and Llama2-7B are similar, Mistral-7B significantly outperforms Llama2-7B. This could suggest that Mistral is utilizing more tokens or scoring them with finer granularity, which results in fewer redundant tokens and thus makes eviction less effective.
{
Furthermore, it is possible that Mistral’s use of grouped-query attention (GQA), which inherently involves compression, may make it more challenging to increase the eviction ratio effectively in this context.}

\begin{table*}[h]
    \centering
    \caption{Performance comparison between Mistral-7B and Ours across various tasks.}
    \label{tab:mistral}
    \small
    \centering
    \begin{tabular}{cccccccccc}
        \toprule
        ~ & Metric & PIQA & ARC-C & ARC-E & RTE & COPA & BoolQ & HellaSwag & Avg. \\
        \midrule
        Mistral-7B & Acc. & 80.09 & 42.37 & 63.14 & 48.01 & 76 & 64.22 & 73.02 & 63.84 \\
        \midrule
        \multirow{2}{*}{Ours} & Acc. & 75.90 & 34.24 & 55.2 & 48.01 & 65 & 62.2 & 67.91 & 58.35 \\
        ~ & Eviction & 37.14 & 39.48 & 37.80 & 40.93 & 45.27 & 44.68 & 50.92 & 42.32 \\
        \bottomrule
    \end{tabular}
\end{table*}

\label{sec:mistral}

\subsection{More Visualization}

\label{sec:more_visual}

{
\Cref{fig:visual2} provides a comprehensive view of the layers and attention heads from \Cref{fig:visual}.
}

\begin{figure}[h]
    \centering
    \includegraphics[width=\linewidth]{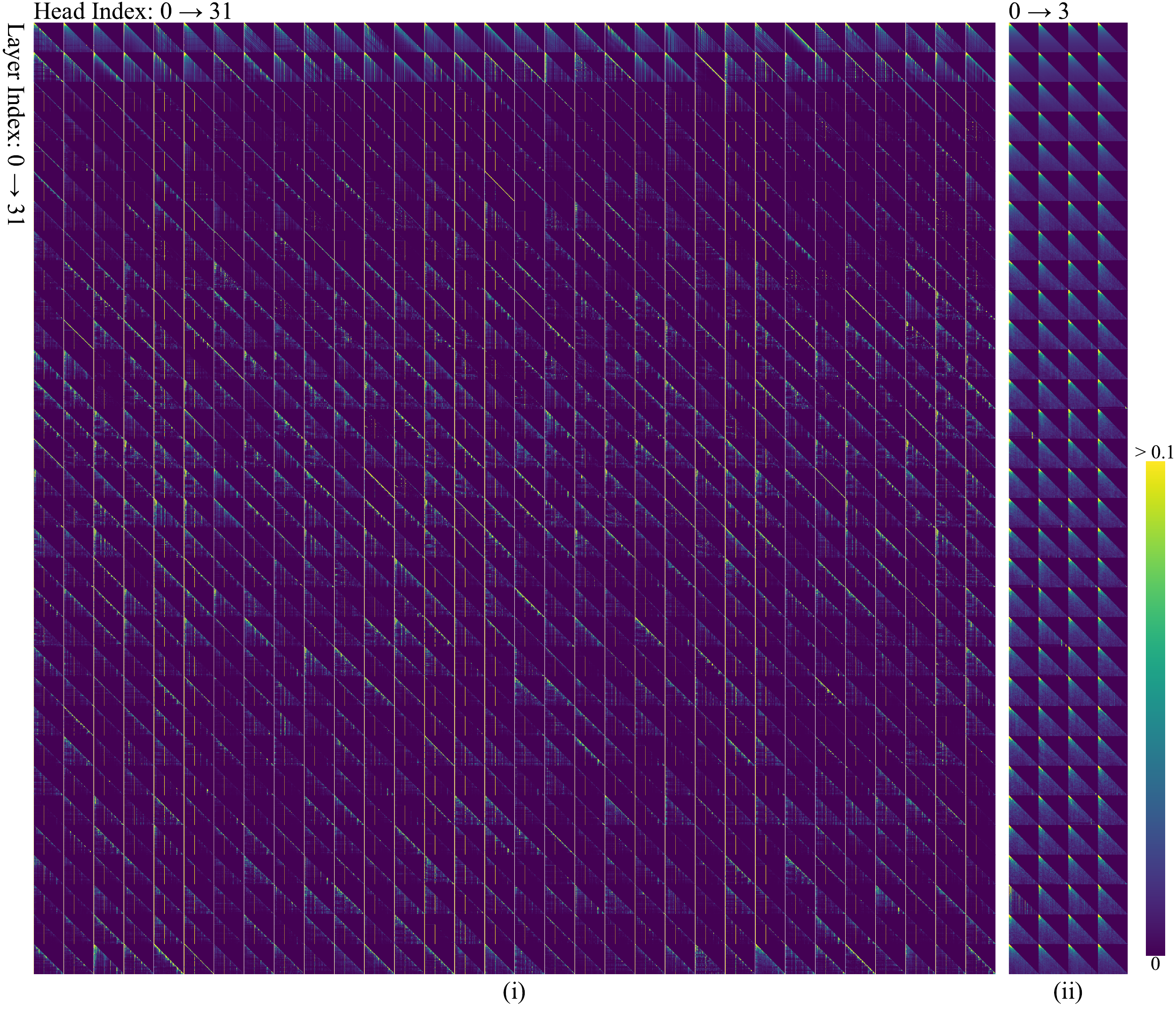}
    \caption{The complete version of \Cref{fig:visual}.
    Two key observations emerge in (i):
    \hlgray{1.} the first two layers are denser compared to the subsequent layers, and
    \hlgray{2.} bright-yellow vertical lines, representing critical tokens for inference, consistently appear across heads in deeper layers.}
    \label{fig:visual2}
\end{figure}

\subsection{More Ablation}

{
In this section, we present additional ablation results in \Cref{tab:comparison2}.}

\label{sec:more_ablation}

\begin{table*}[h]
    \centering
    \caption{
    Exploring the impact of the number of recent tokens, viewed from the perspective of the attention matrix and considering slanted retention patterns.
    (1) corresponds to the setup described in \Cref{sec:sft}, where only the current token is retained, and thus reflecting only the diagonal retention in the attention matrix.
    For (6-1) and (6-2), the number of recent tokens retained is set to 5 and 10, respectively.
    The results suggest that increasing the number of recent tokens does not necessarily enhance performance under the AG framework.
    Further exploration of how to manage recent tokens, such as applying learnable weighted strategies, could be an interesting direction for future work.}
    \label{tab:comparison2}
    \small
    \centering
    \begin{tabular}{cccccccccc}
        \toprule
        ~ & Metric & PIQA & ARC-C & RTE & COPA & BoolQ & OpenBookQA & Avg. \\
        \midrule
        \multirow{2}{*}{(1)} & Acc. & 82.15 & 59.66 & 64.26 & 93.00 & 86.82 & 78.80 & 77.45 \\
        ~ & \%Evict. & 66.16 & 48.31 & 65.47 & 45.40 & 67.46 & 67.17 & 60.00 \\
        \midrule
        \multirow{2}{*}{(6-1)} & Acc. & 83.08 & 50.85 & 65.34 & 82.00 & 87.31 & 73.20 & 73.63 \\
        ~ & \%Evict. & 65.15 & 40.96 & 64.29 & 21.37 & 67.49 & 63.69 & 53.83 \\
        \multirow{2}{*}{(6-2)} & Acc. & 81.61 & 53.56 & 60.29 & 82.00 & 87.37 & 74.20 & 73.17 \\
        ~ & \%Evict. & 65.66 & 44.48 & 65.14 & 24.28 & 68.18 & 63.44 & 55.20 \\
        \bottomrule
    \end{tabular}
\end{table*}

\end{document}